# Stable diffusion models reveal a persisting human–AI gap in visual creativity

S. Rondini[1,2], C. Alvarez-Martin[1,3], P. Angermair-Barkai[4,5], O. Penacchio[2,6], M. Paz[2], M. Pelowski[4,5], D. Dediu[7,8,9], A. Rodriguez-Fornells[1,3,9,10,11*†], X. Cerda-Company[2,6*†]

[1] Cognition and Brain Plasticity Unit, Bellvitge Biomedical Research Institute, L'Hospitalet de Llobregat, Spain
[2] Bridging AI and Neuroscience, Computer Vision Center, Bellaterra, Spain
[3] Department of Cognition, Development and Educational Psychology, University of Barcelona, Barcelona, Spain
[4] Vienna Cognitive Science Hub, Vienna, Austria
[5] Faculty of Psychology, University of Vienna, Vienna, Austria
[6] Computer Science Department, Universitat Autonoma de Barcelona, Bellaterra, Spain.
[7] University of Barcelona Institute for Complex Systems (UBICS), Barcelona, Spain
[8] Department of Catalan Philology and General Linguistics, University of Barcelona, Barcelona, Spain
[9] Catalan Institution for Research and Advanced Studies (ICREA), Barcelona, Spain
[10] Aix-Marseille University, Iméra, Marseille, F-13000, France
[11] Institute of Neurosciences (UBNeuro), University of Barcelona, Barcelona, Spain

* Corresponding authors
[†] Equal senior contribution

# Abstract

While recent research suggests Large Language Models match human creative performance in divergent thinking tasks, visual creativity remains underexplored. This study compared image generation in human participants (Visual Artists and Non-Artists) and using an image-generation AI model (two prompting conditions with varying human input: high for Human-Inspired, low for Self-Guided). Human raters (N = 255) and GPT-4o evaluated the creativity of the resulting images. We found a clear creativity gradient: Visual Artists > Non-Artists ≥ Human-Inspired GenAI > Self-Guided GenAI. Increased human guidance strongly improved GenAI's creative output, bringing its productions close to those of Non-Artists. Notably, human and AI raters also showed vastly different creativity judgment patterns. These results suggest that, in contrast to language-centered tasks, GenAI models may face unique challenges in visual domains, where creativity depends on perceptual nuance and contextual sensitivity— distinctly human capacities that may not be readily transferable from language models.

# Introduction

The recent popularization of Generative AI (GenAI) models, largely successful at mimicking human verbal and visual productions (*1*), have stimulated a growing body of research focused on their creative potential (*2*). GenAI models have been dubbed "Artificial Muses" (*3*), "Engines of Wow" (*4*), "Socio-Technical Monsters" (*5*), depending on the observer's placement on the AI utopia-dystopia spectrum (*6*). Large Language Models (LLMs) are GenAI systems designed for natural language processing and generation, typically pre-trained on vast text corpora (*7*). In the field of computational creativity, this class of models has now been evaluated largely through divergent thinking (DT) assessments (*8*–*10*). DT is generally understood as the ability to form associations between semantically distant ideas, breaking away from prevailing modes of thought, leading to original solutions to open-ended problems, with DT productions usually being assessed across dimensions of Flexibility, Fluency, Originality and Elaboration (*11*, *12*).

The existing research has repeatedly shown that LLMs may often outperform humans in DT (*2*, *13*, *14*), with the exception of a few studies where the most creative humans still outdid the models (*3*, *10*, *15*). Furthermore, when it comes to creativity *perception*, LLMs have been employed to evaluate both human and GenAI DT tests' responses and have been proven to provide accurate scores, with high correlation with the creativity judgements of human raters (*3*, *16*). Overall, considering these recent studies and the standardized definition of creativity as the process leading to the production of something both novel and useful (*17*), GenAI could indeed be regarded as a 'creative agent,' and a remarkably powerful one at that. Nonetheless, points of criticism around computational creativity have also emerged, from GenAI's reliance on statistical recombination of training data in its creative process to, most importantly, the issue of prompt-dependency, arguably leading to a lack of true creative agency and intentionality in the models (*3*, *18*, *19*).

One other existing gap in our understanding of GenAI involves visual creativity, and especially the idea generation or imaginative process. While the vast majority of computational creativity studies have focused on language-based DT tasks in LLMs, visual creativity in image-generation models remains largely overlooked. A series of studies on one-shot image generation across different GenAI models has evaluated their ability to mimic human sketch drawing without, however, specifically addressing creativity itself (*20*–*22*). Another study examined GPT-4o's creative interpretation of abstract visual stimuli, however only analyzing the model's *verbal* outputs according to divergent thinking parameters (*23*). Overall, the unilateral focus of previous studies on language-based GenAI and DT as the main dimension of creativity, make any broad claims regarding the creative advantages in computational agents premature.

Visual creativity, specifically in the form of visual imagery generation or imagination as an end in itself, is often regarded as a key counterpoint to its verbal dimensions (*24*). From an evolutionary point of view, imagination has been linked to the inception of abstract thought, self-idealisation and consciousness as a whole, while also acting as a bridge between the sensory and cognitive dimensions, allowing us to mentally explore possible scenarios in the absence of direct sensorial stimuli (*25–27*). Visual mental imagery (VMI), or "seeing with the mind's eye" (*28*, *29*), engages neural networks spanning from perceptual vision, including early visual areas, to higher-order elements from episodic and autobiographical memory (*29–32*). What we term "Creative Mental Imagery" (CMI) in this paper, however, refers to the internal creation of original images: a testament to the brain's capacity to synthesize new ideas and scenarios independent of direct perception. CMI has been shown to support our information-processing through depictive representations (*31*), participating in our evolution as ever-innovating, ever-creative beings (*33*). Despite the theory of DT having established itself as the dominant paradigm in creativity research (*34*), the involvement of mental imagery in creative processes is substantial (*35–38*).

The embedded nature of visual imagination, spanning from perceptual to higher-order, semantic abilities, becomes manifest in the complexity of its product, creative imagery, their scoring based on their physical properties, as well as their conceptual content (*33*). Current DT studies have so far drawn conclusions on GenAI creative superiority based on verbal creative outputs, easily explained by the computational disparities between human and AI agents. CMI thus constitutes a particularly interesting strategy for the investigation of computational creativity from a more complex lens, to include perceptual and semantic skills, beyond the agents' computational abilities.

Thus, our project focused on the visual imagination abilities of a Stable Diffusion model compared to human samples, by observing both production and appraisal of creative mental imagery. Stable Diffusion are a class of diffusion-based text-to-image model, which generate images starting with random pixel noise in an image grid and iteratively refine it, using the text prompt at each step to guide the denoising toward features and patterns that match the prompt until a coherent image forms (*21-22*). At present, they constitute the class that best mimics human sketch-drawing (*22-23*), and was therefore selected for the project. Potential similarities and contrasts in the creative visual behaviour of humans and GenAI may shed light on the important differences in how computational and human agents interact creatively with visual stimuli, both perceptually and semantically, thus offering a broader stance on creativity in state-of-the-art models. Additionally, we address the underlying issue of intentionality and agency in the creative process of GenAI models, by introducing different degrees of human guidance through prompting, aiming to determine the extent to which human input influences GenAI's successful creative outputs.

The present study comprised four phases (for a scheme of the study, see Figure 1 and Materials and Methods for details). Phases I and II were based on the Test of Creative Imagery Abilities (TCIA) (*34*), and aimed at generating two distinct datasets of creative drawings: by humans, comprising visual artists (N= 27) and non-expert participants (N = 26), and from GenAI models. Since previous studies on one-shot image generation had determined Diffusion models to be the most proficient class of GenAI models at human drawing reproduction (*20-22*), we selected a Stable Diffusion model for our task (39).

In the TCIA, participants were presented with a series of 12 abstract stimuli, which they used to mentally generate images. Each participant then listed the images they had imagined and selected one to draw on paper. The TCIA was reproduced in the GenAI groups by employing the TCIA abstract stimuli as ControlNet and a basic prompt to generate drawings. As a key criticism of previous GenAI creativity research had targeted GenAI's lack of agency as pre-emptive of a genuine creative process *(3, 18, 19)*, different levels of prompt elaboration, and thus human guidance, were employed in the two GenAI groups of phase II.

For the Human-Inspired GenAI condition a more specific, concrete prompt containing one of the human-generated ideas from phase I was employed, while the Self-Guided GenAI condition employed the same base prompt with the exception of the added human idea. Finally, phases III and IV focused on the scoring of the total generated dataset of 1,000 images, obtaining separate creativity ratings from both human and GenAI raters through an image rating task using five creativity dimensions (Liking, Vividness, Originality, Aesthetics and Curiosity) (34).

In phase III, the image rating task was carried out online by human raters (N = 255). Prior to the image rating task, raters completed the Aesthetic Responsiveness to Art (AreA) questionnaire (*40*), in order to determine whether differences in aesthetic tendencies led to variations in creativity perception.

Each rater received a unique subset of 120 drawings, comprising an even distribution of images from all four experimental groups, while also ensuring that each image would gather a minimum of 30 ratings. No information was provided on the origin of the drawings. In phase IV, the same rating task was reproduced through the GPT-4o API (*41*).

Due to the novelty of the study, our predictions were exploratory in nature. On one side, assuming that GenAI's linguistic competence generalizes to visual creativity, patterns similar to those of DT studies may be expected, with GenAI agents outperforming human participants (*2*, *13*, *14*), and with the standing possibility of the most creative humans to still place above the models (*3*, *10*, *15*). Furthermore, based on previous findings showing refined prompt-engineering techniques to successfully direct and enhance the creativity of GenAI outputs (*42–45*) XXXX (Nath Paper), it is likely that the GenAI group with increased human prompting will display higher creativity compared to the low-guidance GenAI group.

When it comes to the appraisal of creative images, GenAI performance compared to humans remains relatively understudied, the limited existing research showing weak aesthetic perception competence (*46*). While some models do reach close correlations with human raters, the tasks usually focus on physical attributes of the images, rather than deeper dimensions of creativity (*47–49*). Based on this, stronger similarities between human and GenAI raters may be expected to emerge for the more objective, computable creativity dimensions, with potentially more variation found for the more contextual, subject-dependent features of a creative image.

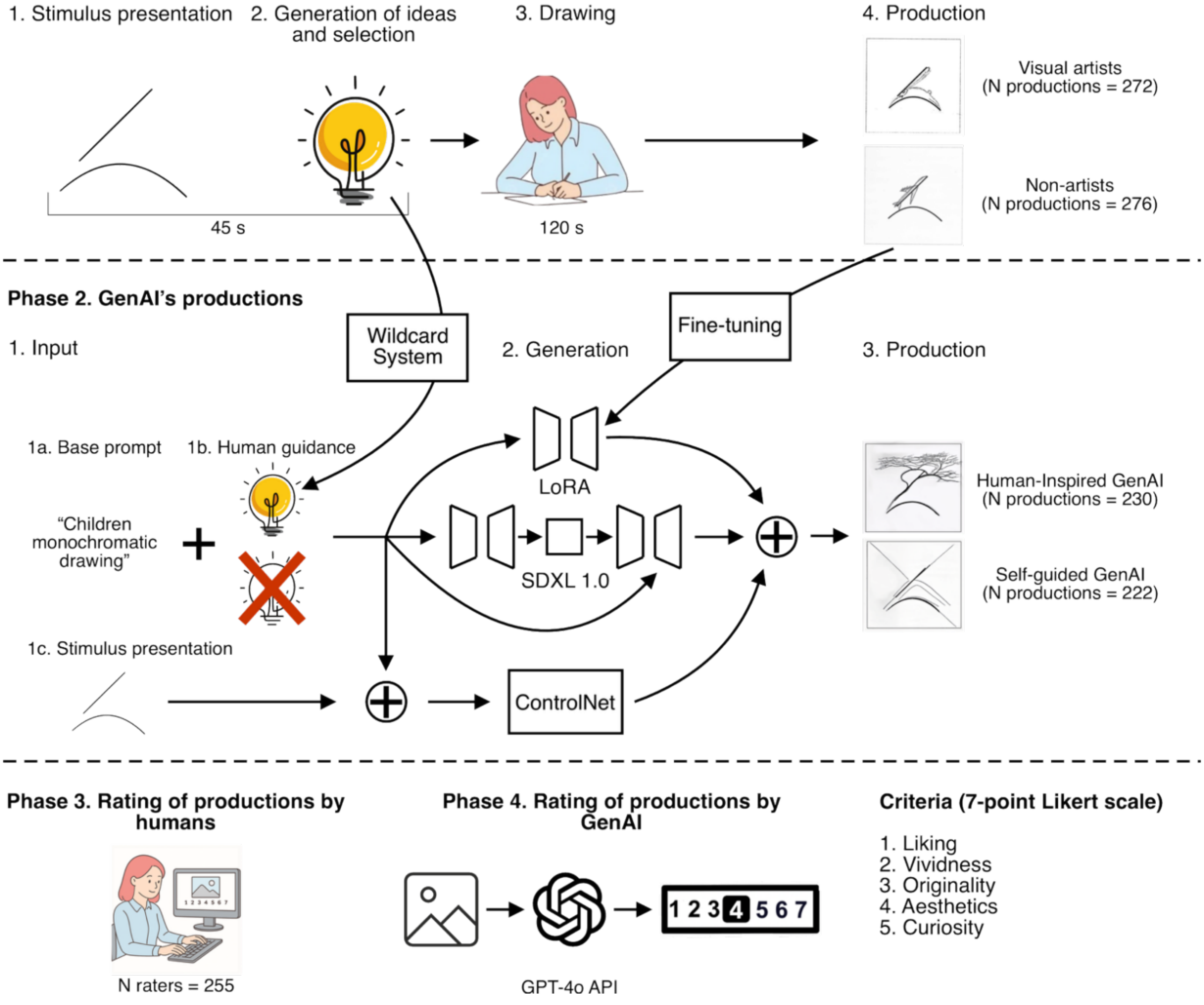

**Figure 1.** Scheme of the study's four phases. **Phase I**: Creative image generation by Visual-Artists and Non-Artists. The resulting human images were used to fine-tune the Diffusion model (SDXL) through Low-Rank Adaptation, while the human-generated ideas were used in the prompt of the Human-Inspired GenAI group. **Phase II**: Creative image generation by GenAI (Human-Inspired and Self-Guided). **Phase III**: Creative image dataset rating by human raters. **Phase IV**: Creative image dataset rating by GPT-4o.

# Results

Examples of the drawing outputs can be seen in Figure 2. The creativity ratings of the image dataset generated in phases I and II were obtained from human and GenAI raters in phases III and IV, respectively. These two sets of creativity ratings were analyzed, first, individually and then compared as follows.

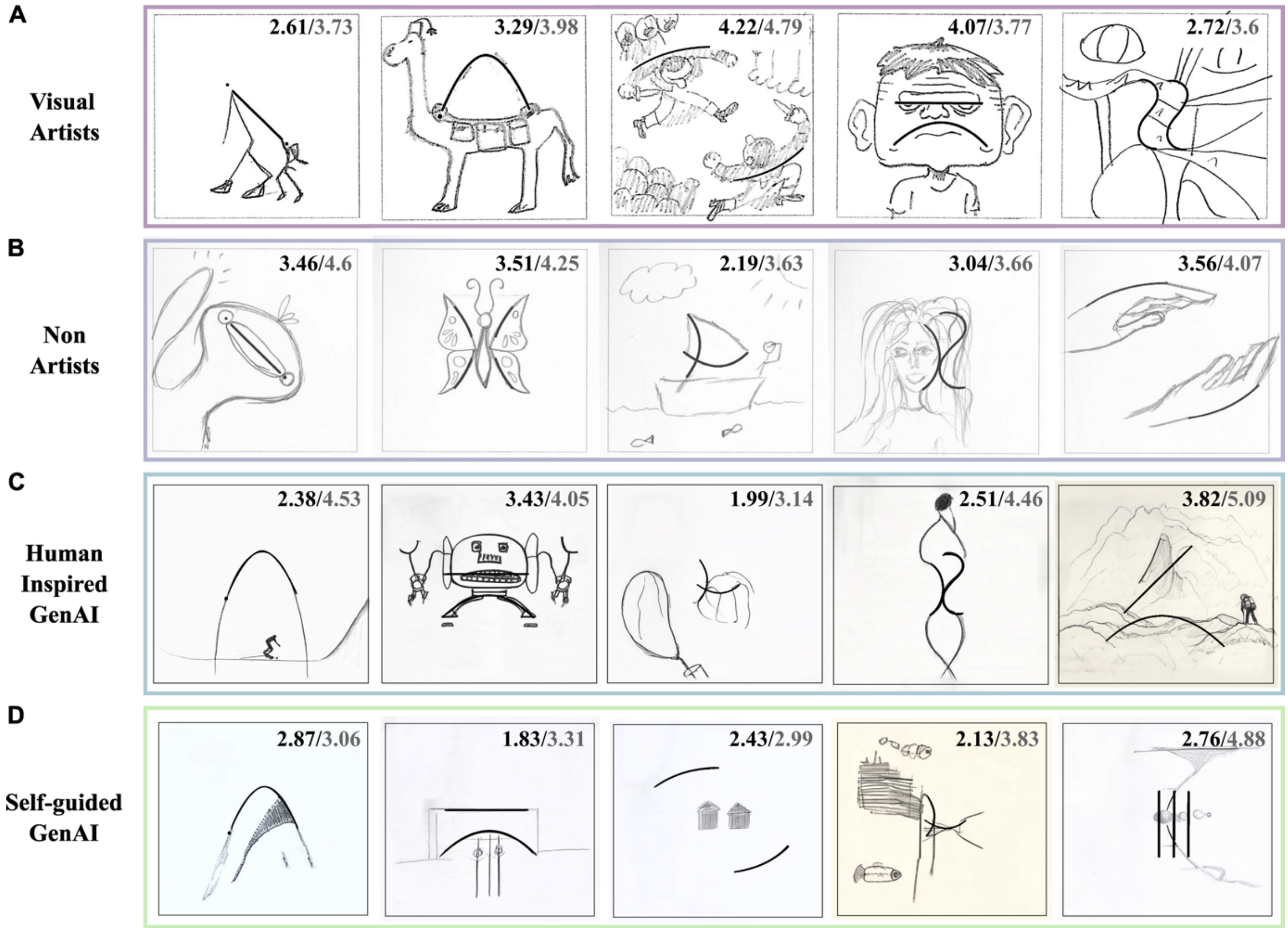

**Figure 2.** Examples of drawings from each category (in rows): Visual Artists, Non-Artists, HI-GenAI, SG-GenAI. The values reported in the right-hand top corner of each drawing corresponds to the drawing's Creativity score by humans and GPT-4o raters, respectively.

## Human ratings: analysis results

*Overall Creativity Analysis*

When studying human ratings, an overall Creativity score was generated from the five dimensions scored by participants (i.e., Liking, Vividness, Originality, Aesthetics, and Curiosity & Interest) using Factor Analysis (FA, see Materials and Methods for more details). An analysis dividing the sample between Human and GenAI images showed that, indeed, human-generated images were scored significantly higher than GenAI ones ($\chi^2(1) = 229.71$, $p < .001$, see Table S1 for model details).

Subsequent analysis through linear mixed-effects models (See Materials and Methods for models' details) revealed image category to have an effect on the overall Creativity score ($\chi^2(3)=457.03$, $p < .001$, see Table S2 for model details). Specifically, the Visual Artists group obtained the highest Creativity scores, followed by Non-Artists images and Human-Inspired GenAI, and lastly, Self-Guided GenAI images (see Figure 3A).

Pairwise comparisons showed overall Creativity scores to be significantly different ($p < .05$) between categories (See Table S2: between Visual Artists and Non-Artists (estimate = 0.42, $p < .001$), between Visual Artists and HI-GenAI (estimate = 0.51, $p < .001$), and between Visual Artists and SG-GenAI (estimate = 0.97, $p < .001$). Importantly, the difference between Non-Artists and HI-GenAI groups was the only non-significant one (estimate = 0.09, $p = .198$), with further complexities emerging when looking at individual dimensions of creativity (see below). This highlighted the marked effect of prompting the GenAI with human ideas on its overall creativity, leading to the model performing on par with non-expert humans.

*Creativity Dimensions Analysis*

The results from linear mixed-effects models showed the same clear descending pattern between categories across all individual dimensions of Creativity (see Tables S3 to S7). As reflected in Figure 3B, pairwise comparisons between categories were significant for all dimensions, with the exception of the comparison between Non-Artists and Human-Inspired GenAI in the Liking and Vividness dimensions.

In summary, these results indicate that images drawn by humans were rated as overall significantly more creative than those generated by GenAI. Within the GenAI group, incorporating human ideas into the prompt led to markedly higher perceived creativity ratings, matching the scores of non-expert humans.

**GPT-4o ratings: analysis results**

*Overall Creativity Analysis*

To assess the effect of image category on the creativity scores in GenAI-rated images, the same analysis performed on human ratings was repeated using the GPT-4o ratings (see Figure 3C). Results from the linear mixed effects model suggested that, although the effect of category was significant ($\chi^2(3) = 43.5$, $p < .001$, see model summary in Table S8), this effect was driven by the Self-Guided GenAI category, which showed significantly lower scores than all other categories. Indeed, analysis comparing between Human and GenAI images showed no significant effect of group ($\chi^2(1) = 2.94$, $p = .09$).

*Creativity Dimensions Analysis*

As done for human ratings, linear mixed-effects models were used to study the effect of image category in the rest of the dimensions rated by GPT-4o (Supplementary Tables S9 to S13). Contrary to human ratings, GenAI ratings did not show a consistent pattern across all Creativity dimensions (see Figure 3D), although Self-Guided GenAI rated significantly lower than the other categories in all dimensions. Interestingly, Human-Inspired GenAI images rated significantly higher in Originality and Curiosity, while the higher Vividness scores were found in the Non-Artists category.

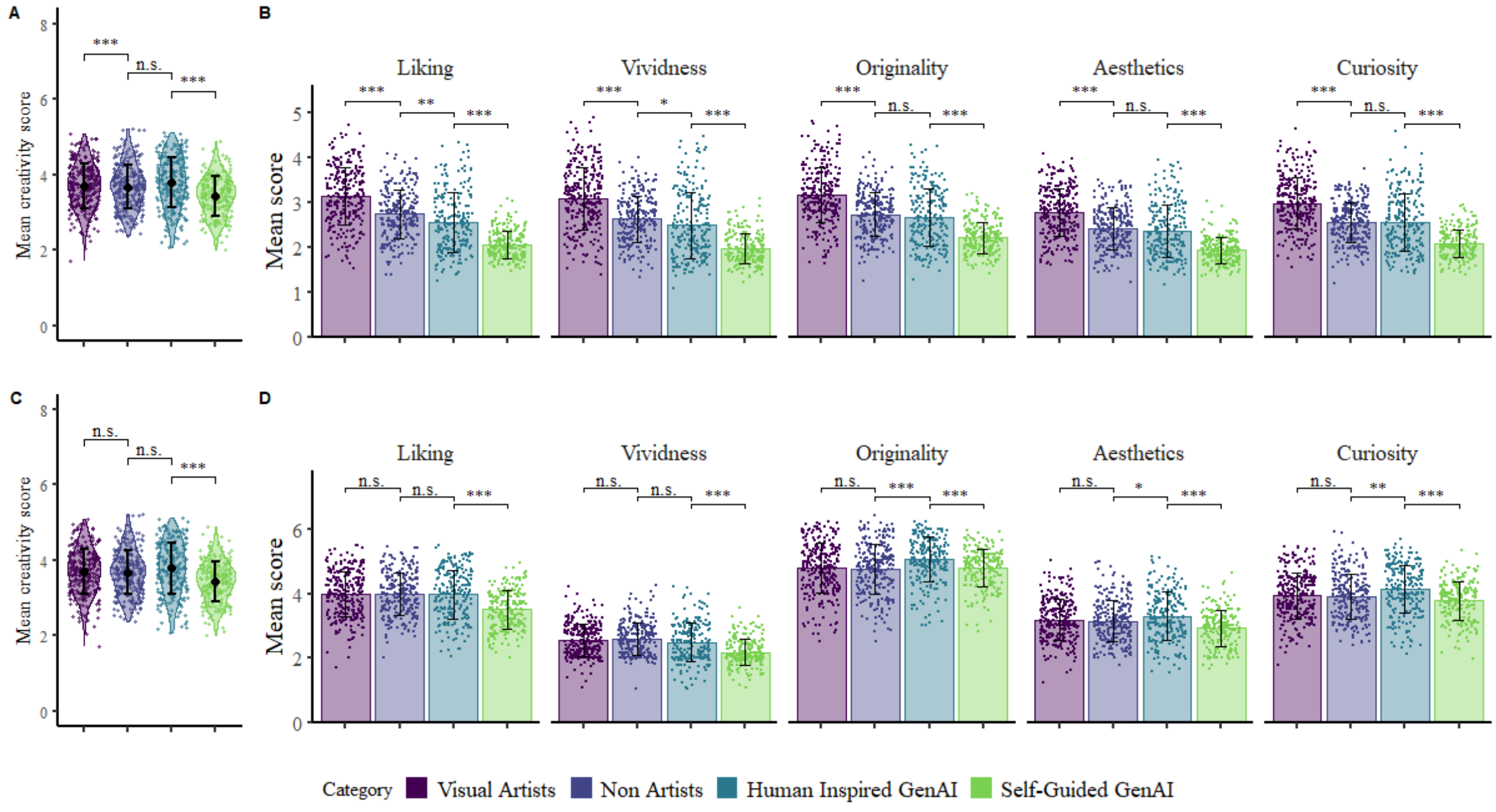

**Figure 3.** Creativity Ratings by Humans and GPT-4o. **A.** Mean Creativity score (1-7) for each image category (Visual Artists, Non-Artists, HI-GenAI and SG-GenAI). **B.** Mean ratings for each Creativity dimension (Liking, Vividness, Originality, Aesthetics, Curiosity) by image category. **C.** Mean Creativity score for each image category. **D.** Mean ratings for each Creativity dimension by image category. Individual points represent image means and error bars represent category mean ± standard deviation. Pairwise comparisons from linear mixed-effect models are shown for adjacent categories. *$p < 0.05$, **$p < 0.01$, ***$p < 0.001$. n.s. = not significant.

### Differences between Human and GPT-4o Creativity ratings

Given the different patterns found for the Creativity rating of human versus GenAI drawings, a new linear mixed-effects model was fitted to test whether these differences were significant (see Table S14). Not only were the ratings from GPT-4o significantly higher than human ratings across all categories ($\chi^2(1) = 621.47$, $p < .001$), but the interaction between image category and type of rater was significant ($\chi^2(3) = 1035.26$, $p < .001$), indicating different ratings patterns. Specifically, while human raters assigned the highest creativity scores to images produced by Visual Artists and rated the other three categories (Non-Artists, HI-GenAI, and SG-GenAI) progressively lower, GPT-4o exhibited a markedly different pattern (see Figure 4). GPT-4o ratings showed increased perceived creativity across all image categories, while displaying less discrimination between them, in some cases even favoring GenAI images. The most pronounced differences between human and GPT-4o ratings emerged for the HI-GenAI and SG-GenAI images. Whereas human raters assigned these categories the lowest overall Creativity scores, GPT-4o rated them at levels comparable to (or even higher than) those given to human-generated images. These patterns highlighted a tendency in the GPT-4o rater to assign more generous creativity scores, particularly in the case of GenAI images, strongly contrasting with the human ratings, which were overall more conservative and showed significant discrimination between image categories.

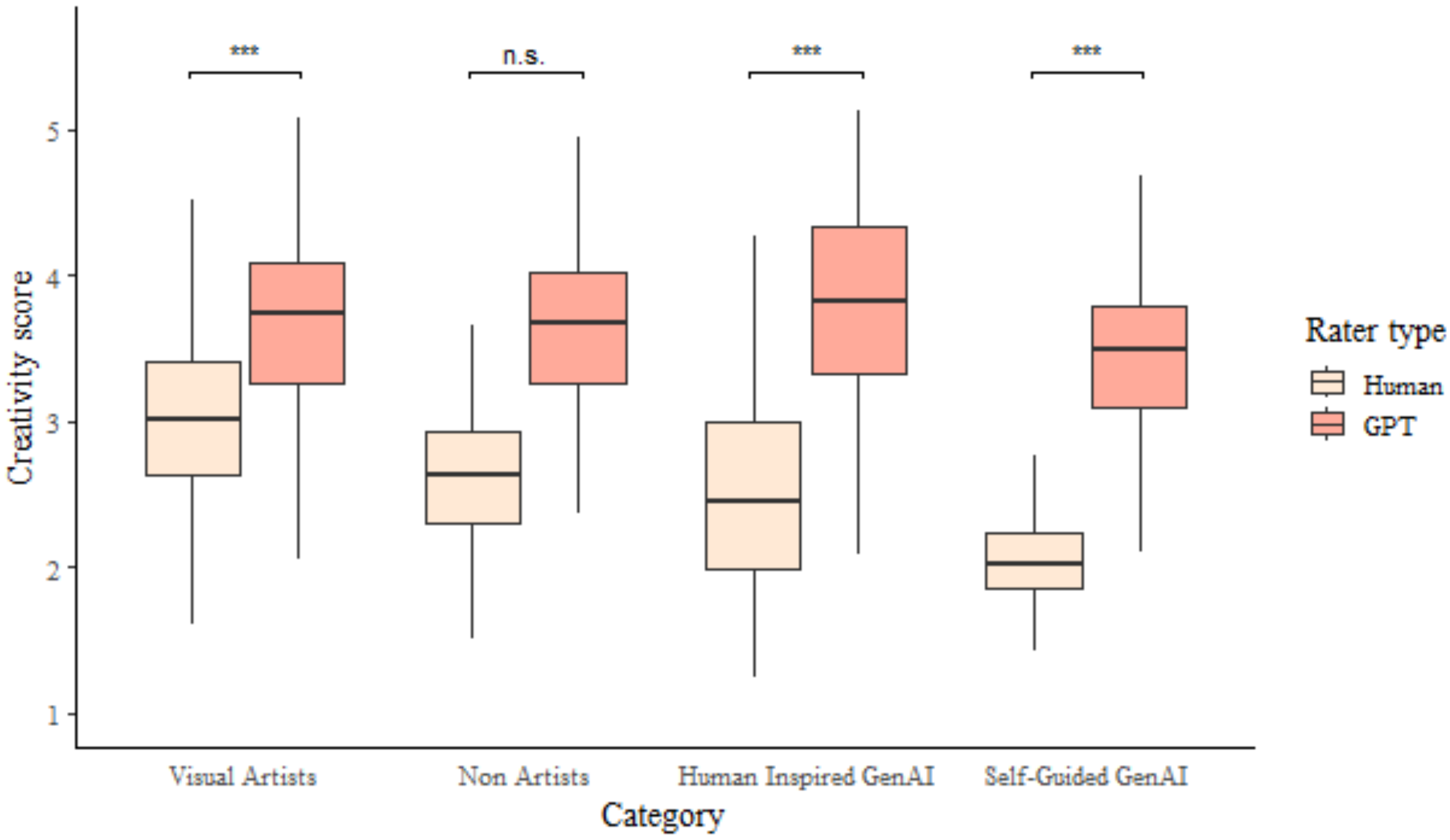

**Figure 4.** Boxplot showing Human vs GPT-4o overall Creativity ratings across image categories (Visual Artists, Non-Artists, HI-GenAI and SG-GenAI).

Moreover, the distribution of the responses on the five dimensions (see Supplementary Figure S1) already hints that human and GPT-4o may be fundamentally different, which is further confirmed by their k-means clustering. As the gap and the silhouette methods suggest (between 2 and 6 clusters), we systematically analyzed all of these, which show that the main division ($k$ = 2; see Figure 5A) is between a cluster of mostly human ratings (34% vs 8%) and another of mostly GPT-4o ratings (41% vs 16%), while higher $k$ values further refine these two clusters (see all details on OSF project).

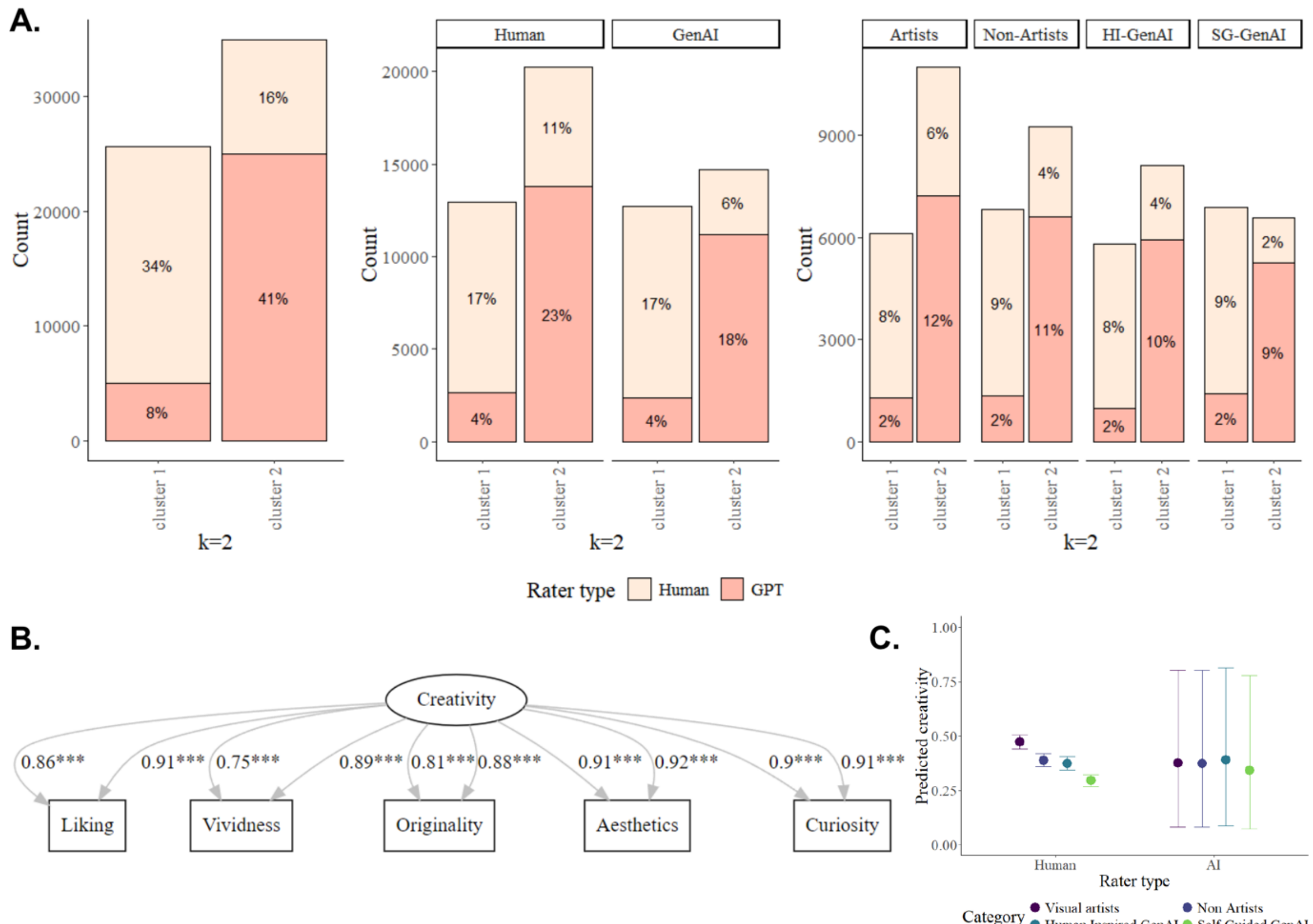

**Figure 5**. **A.** Distribution of the responses among the k-means with k=2 clusters. Left: by rater type (colour, human vs GPT-4o). Middle: by rater type (colour) and drawer type (human or Stable Diffusion). Right: by rater type (colour) and image category (Non-Artists, Visual Artists, Human-Inspired GenAI and Self-guided GenAI). **B.** The one-factor model fitted to the whole dataset, split by rater type (human versus GPT-4o). There are two arrows and loadings for each indicator, the first being for the human raters and the second for the GPT-4o rater. The loadings are standardized, and the stars represent their significance level (here, all $p < .001$). **C.** Predicted Creativity score from a beta regression on the interaction between rater type and image category while controlling (as random effects) for item and individual rater.

The five dimensions result in 8.8% Guttman errors bigger than (Q3 + 1.5*IQR) = 10 and have a set homogeneity value $H$ = 0.815 with 95% confidence interval (0.812, 0.818) which is significantly higher than the recommended 0.30, suggesting that the scale is homogeneous; this is further supported by the fact that all items have a homogeneity ≥ 0.725. Likewise, Automatic Item Selection Procedure (*aisp*) finds that all dimensions belong to the same scale for all $c$ between 0.05 and 0.60. Principal Component Analysis (PCA) on human ratings found that the first principal component explains 84.9% of the variance (see Supplementary Figure S2A) and all five dimensions have loadings of the same sign and comparable strengths (see Supplementary Figure S2B), while the same analysis on GPT-4o ratings revealed a first component that explains 78.3% of the variance (see Supplementary Figure S2C), with more variance between the strengths of the loadings (see Supplementary Figure S2D). These results are supported by the exploratory Factor Analysis (EFA) with a one-factor model explaining 76.6% of the variance when all judges are pooled together, but 81.8% when only human judges are considered, and 71.9% for GPT-4o. Fitting a one-factor confirmatory multi-group (by judge type) Factor Analysis (CFA) model (see Figure 5B) results in a significant *p*-value ($\chi^2(10)$ = 23,815.2; $p < .001$), but with decent fit indices (CFI = 0.92, TLI = 0.84 and NNFI=0.84), which is not surprising given the amount of data and the complexity of the design. Interestingly, checking its configurational, weak and strong invariance, strongly suggests that both the factor loadings ($\chi^2(4)$ = 3,515; $p < .001$) and the latent means ($\chi^2(4)$ = 50,567; $p < .001$) are different between human and GPT-4o ratings. From this model we estimated the latent "Creativity score" (using the loadings for humans and GPT-4o, separately), which is our outcome of interest in the following regression models.

While the rater type (human vs GPT-4o) did not have a significant main effect ($p = .94$), it did significantly interact with the drawer type ($p < .001$) and the category ($p < .001$; see Figure 5C). That is, (a) while the humans made a clear distinction between kinds of drawings, with a clear hierarchy Visual Artists ≫ Non-Artists ≳ Human-Inspired GenAI ≫ Self-Guided GenAI, GPT-4o did not, and (b) while the error bars of the human raters were very constrained, those for GPT-4o were notably wider. Taken together, these findings suggest that while *on average* human and GPT-4o scores are similar, this may be misleading as the human scores are rather focused and in agreement with each other for a given image category, while GPT-4o's are scattered across the whole range of possible scores.

The visual stimulus used to prompt a drawing had a main effect and interacted significantly with the rater type. However, in practice, this resulted in minimal changes in the scores that did not affect the qualitative results (See Table S15 for model details). Focusing on the human raters, individuals were divided into Low and High AreA groups using a median split (AreA = 2.93). The main factor influencing their evaluations was their own AReA score. A significant positive main effect was found ($p < .001$; see Figure 6 and Table S16 for model details), indicating that individuals with higher AReA scores reported higher Creativity ratings than those in the Low AReA group. However, in this case the interaction is not significant, the only difference in the rating pattern being a significant difference between the Non-Artists and Human-Inspired GenAI in the High AReA group that disappears in the low group.

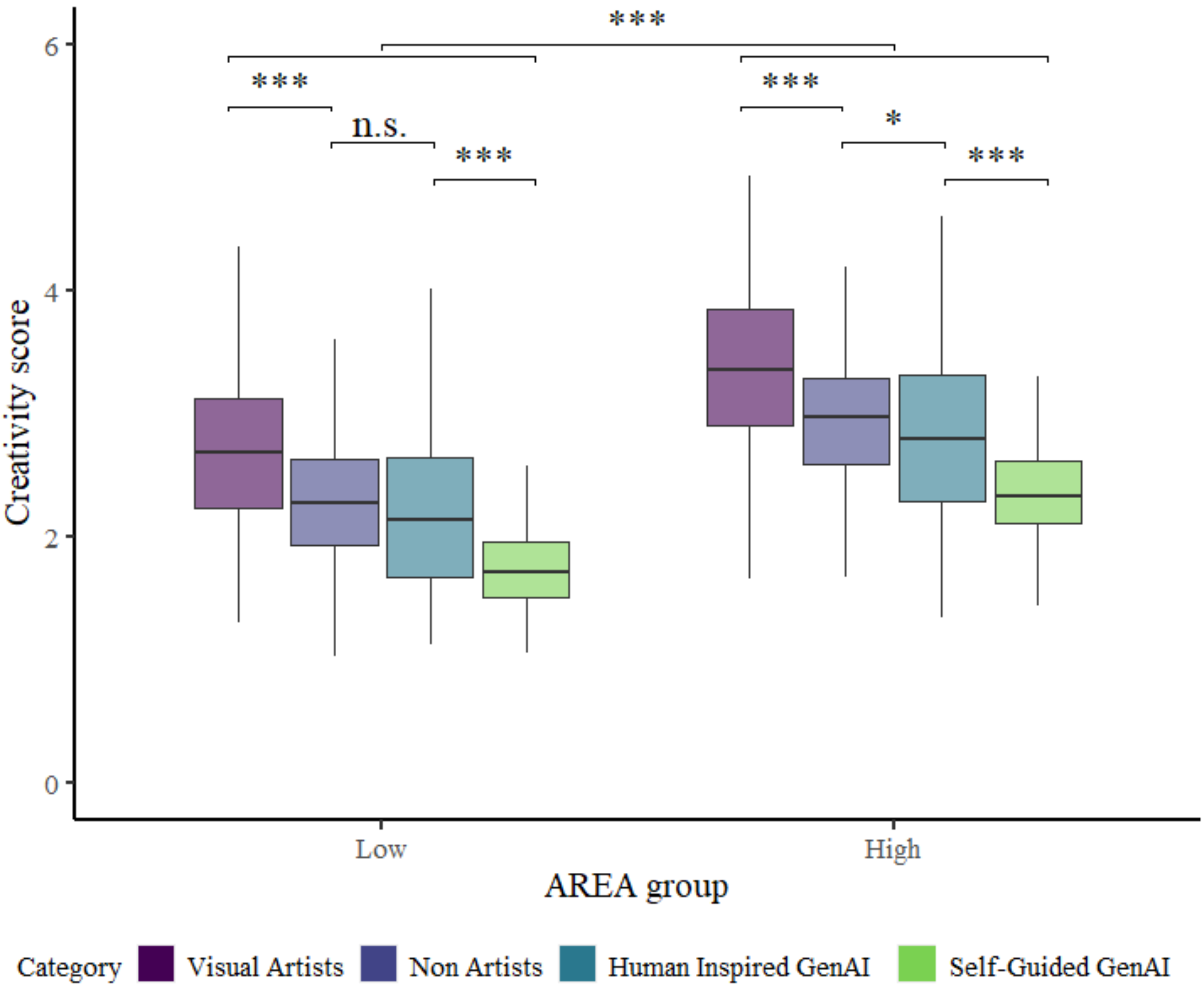

**Figure 6.** Impact of AReA scores on Creativity ratings. High-AReA participants assign overall higher creativity ratings than low-AReA participants. The pattern of discrimination between image categories remains constant.

## Discussion

The study explored visual creativity in image-generation models and humans through the Test of Creative Imagery Abilities (TCIA). To allow comparison with a larger range of human performance at the TCIA, a group including both Visual Artists and Non-Artists was recruited. Furthermore, in order to investigate issues of prompt-dependency and their impact on computational agents' creative performance, different degrees of human guidance through prompt-elaboration (HI-GenAI and SG-GenAI) were introduced. Lastly, the creativity of human and GenAI images was scored by both human and AI raters.

The results showed a clear human advantage at the TCIA, with the Visual Artists' productions being rated as the overall most creative, followed by the Non-Artists' and HI-GenAI with comparable scores and, lastly, SG-GenAI (see Figure 3A). Although the GenAI groups generally performed considerably worse than the human groups, providing a more elaborate, human-inspired prompt elevated the HI-GenAI group's outputs to human-level performance. Lastly, when comparing creativity ratings, GPT-4o failed to mirror humans in their creativity appraisal: Neither overall perceived creativity scores, nor image category discrimination matched between human and GenAI ratings (see Figure 4).

The major differences observed in GenAI visual creativity performance in our study stand in stark contrast with existing research, where GenAI consistently surpassed humans in verbal divergent thinking (DT) tasks. In previous computational creativity studies, generative algorithms generally performed well on Flexibility, Elaboration and Fluency, with sometimes lacking or average results in Originality (*3*, *15*). While these results make intuitive sense due to GenAI's natural computational advantages in the generation of high numbers of elaborate, semantically flexible ideas, they also highlight core issues of employing DT tests as principal computational creativity measure. By overly focusing on creativity dimensions such as Fluency, Flexibility and Elaboration, heavily dependent on computational power and training-data size, the obvious inequalities between human and GenAI agents lead to the conclusion that computational agents outperform human creativity.

The highly specific structure of the queries in DT tests may constitute a further advantage for GenAI agents: clear and constrained prompting is a major factor in the optimal functioning of LLMs, while these same conditions may not be as encouraging of human creativity (*34*, *50*). Current architectures, functioning on closed-world formalisms, excel at finding optimal solutions to well-defined problems, while still, however, struggling when faced with the uncertainty constituted by the emergent possibilities and outcomes of open-world problems (*51*). Additionally, previous research has shown that humans tend to achieve higher Originality scores in figural DT tests compared to verbal ones (*50*). Figural DT tasks tend to bypass the obvious, overlearned associations favored by verbal tasks, resulting in lower Fluency but higher Originality (*50*). The structure of figural DT tests strongly resembles that of the TCIA, based entirely on abstract, non-semantic stimuli. Accordingly, the open-ended nature of our task could partially explain why, in our study, human participants outperformed GenAI across all creativity dimensions—strongly contrasting with previous studies relying on verbal DT tasks, where specific, semantically loaded prompts favored GenAI performance, while potentially constraining human creativity.

On the other hand, the inclusion of a human-generated idea in an otherwise basic, abstract prompt, succeeded in elevating the creative results of the HI-GenAI group, nearing the performance level of human Non-Artists. Previous studies have argued that despite its creative potential, GenAI remains restricted as it does not experience the spontaneous need to ideate, to imagine, in the way human cognition does, relying on human input to simulate a real-world connection in which that creative potential can instantiate (*3*, *52*). GenAI agents are likely inherently solipsistic, existing and acting in a “small world” in an entirely predefined and formalized manner, their framing being constituted by the training dataset and the model’s architecture (*51*, *53*, *54*). Unlike human agents, their understanding of the outside world is not built through interaction, but rather through pure probabilistic associations gathered during training (*55*). Human cognition, on the other hand, evolved in the context of maximizing a biological agent’s adaptation and survival in an uncertain world, requiring a high degree of openness and generalization abilities (*51*). Human creativity shares the same strong evolutionary roots, the ability of bringing forth valuable and novel solutions to ever-changing, open-world problems (*56*–*58*). It is thus largely relational, contextual and sensitive to outer world conditions, allowing for flexible, adaptive responses to a complex environment (*51*, *59*). Human creativity, as defined by a novel and useful production, whether emerging through DT or imagination tasks, does not exist in a vacuum, but is rather a process instantiating between an embodied agent in an on-going interaction with its environment (*60*). In the case of GenAI agents, what prompting likely does is to provide the real-world connection they lack to generate a contextually useful, creative product.

In contrast with DT tasks’ queries, highly specific and semantic in nature, the abstract visual stimuli of the TCIA constituted an extremely open-ended, vague framing, which when complemented by equally vague and open-ended prompting, resulted in clearly lacking and uncreative productions in the SG-GenAI group. The TCIA stimulus alone failed to trigger any creative image generation, as it would have in a human agent, who is able to leverage their biological memory and perception, as well as real-world contextual information, to spontaneously and flexibly create (*21*, *61*). The addition of a concrete, human-generated idea in the HI-GenAI group’s prompt essentially simulated in the model the creative process taking place in the human participants when exposed to the TCIA stimuli, considerably improving the perceived creativity of the productions. The performance gap between the SG-GenAI and the HI-GenAI groups resulting from more directed prompting underscores the importance of human guidance for successful GenAI creativity, while also showing that, at present, GenAI models cannot be regarded as autonomous creative agents.

Furthermore, the drastic differences between human and GenAI raters at the image rating task (see Figure 4) strongly contrasted the studies which had found LLMs to constitute accurate raters of verbal DT responses, highly correlating with human ratings (*3*, *16*). Despite the variance in aesthetic tendencies among the sample of human raters (See Figure 5c), the creativity appraisal patterns remained consistent across the different experimental categories, with a clear descending trend from the Visual Artist group to the SG-GenAI group. On the other hand, the GPT-4o ratings showed no clear discrimination between image categories, and inconsistent patterns between the individual creativity dimensions. The dynamics underlying GenAI's issues in reproducing the human creative process may also explain the gap in visual creativity appraisal. Text and images are processed in fundamentally different ways by GenAI models: text is sequential and symbolic, thus easily captured by the probability-based relationships on which transformer architectures rely (*51*, *62*). Specifically, the scoring of DT tests responses by LLMs rests heavily on their number, elaboration and their content's semantic distance, all easily quantifiable and computable dimensions (*63*). On the other hand, images, creative or not, are high-dimensional and spatial, requiring models to understand complex visual structures, leading to additional perceptual challenges arising in the rating of creative TCIA drawings. The complexities of visual perception increase in the appraisal of visual creative products beyond physical features, their creativity being largely determined by the observer's inner psychological state and outer socio-cultural context (*64*). In essence, the fundamentally different perceptual strategies and lack of contextual awareness in LLMs (*21*, *51*, *52*), while not particularly impacting language processing, may play a larger role in creative image evaluation.

Overall, the different patterns emerging from GPT-4o and human ratings solidifies the stance that computational agents' visual creative behavior does not yet match that of humans, its functioning likely relying on evolved biological characteristics and continuous adaptive interaction with the open environment, features that have not, at present, been successfully reproduced in GenAI models.

## Conclusions

The present study probed the current state of GenAI visual creativity by directly contrasting human and GenAI performance in the production and evaluation of creative images. Using a vision-based imagination task, rather than standard verbal tests, we showed that GenAI remains markedly limited in creative capacity, underscoring the need for broader measures to better triangulate the complexity of creativity in computational agents. The challenges GenAI faces in visual creativity could stem from fundamental divergences with human cognition and perception, including a reduced sensitivity to real-world complexity. The improvement observed under human guidance hint at the deeply embedded nature of human creativity, as an evolutionary process at the cusp between a biological agent and the complex, dynamic environment it inhabits, and point to a fundamental limitation in current GenAI. Overall, the models' reliance on human prompting indicates that, while GenAI *can* exhibit creative abilities when properly guided, it does not yet reach autonomous human-level performance.

# Materials and Methods

**Materials**

An adapted version of the TCIA (*34*) was employed, where each participant is presented with a sequence of 12 abstract stimuli. Six out of the 12 stimuli were taken from the original TCIA, while the remaining six were newly created (see Supplementary Figures S3 and S4). For the rating of our dataset, to the original TCIA scoring criteria "Vividness" and "Originality" (*34*), creativity measures of "Liking", "Aesthetics" and "Curiosity" were added (see Section S1 for detailed creativity measures selection).

The studies involving human participants, both in-lab and online, were performed in accordance with the Declaration of Helsinki, and approved by the Clinical Research Ethics Committee of the Bellvitge University Hospital (PR335/21). Participants received monetary compensation, and they provided informed consent for their participation in the study.

**Phase I. Creative Image Generation in Human Artists and Non-Artists**

This in-lab study examined two distinct groups (visual artists and non-artists) to investigate human performance in CMI, and to assess potential differences in creative imagination abilities between individuals who routinely engage in CMI-based activities (Visual Artists) and general population (Non-Artists; see Figure 1). Participants from both groups were economically compensated for their participation. The groups were composed, respectively, of:

1. Human Visual Artists (n = 27; mean age = 25.4 ± 4.8 y.o.; 14 female, 13 male), recruited from local fine arts institutes and ateliers (mean years practicing visual arts = 6.03 ± 3.95 y.o.). Each participant was required to provide proof of enrolment in, or completion of, a visual art-related degree or, alternatively, their professional art portfolio.
2. Human Non-Artists (n = 26; mean age = 25.2 ± 4.04 y.o.; 19 female, 7 male). Participants were recruited via Sona Systems© (Sona Systems Ltd., Tallinn, Estonia), through our faculty.

*Procedure*

The TCIA was administered to human participants using Psychopy (*65*). Participants were not informed of the task's aim, and no mention was made of creativity or imagination. The TCIA stimuli were presented in a randomized order. Each stimulus was shown individually for 45s, and participants were asked to use it as a base to mentally generate any number of possible mental images (Figure 1). After each stimulus presentation, participants listed all the ideas they conceived, before selecting one to reproduce on a provided sheet of paper presenting the stimulus they had just observed. A maximum of two minutes was allocated for the completion of each drawing. A total of 660 images were generated: 324 by the Visual Artist group, and 312 by the Non-Artist group.

**Phase II: Creative Image Generation with GenAI - Stable Diffusion models**

This experiment aimed at reproducing the TCIA in a Stable Diffusion model, enabling a comparison between GenAI creative imagination with humans (See Figure 1). Two sets of GenAI drawings were produced, varying in the degree of human intervention through the elaboration of prompts:

1. Human-Inspired (HI)-GenAI: human ideas generated by the human groups were added to the basic prompt (see below).
2. Self-Guided (SG)-GenAI: no human ideas were added to the basic prompt (see below).

*Procedure*

GenAI images were generated employing a specifically fine-tuned Stable Diffusion XL Base 1.0 (*39*) (See Section S2 for details on fine-tuning and image generation). Each TCIA stimulus (see Supplementary Figure S3 and S4) was employed as ControlNet input, paired with the base prompt "Children monochromatic **[idea]** drawing", where **[idea]** (e.g. "whale", "bridge", "camping tent") was randomly selected from human responses from Phase 1, using a Wildcard system. The SG-GenAI category used the same base prompt structure without the added **[idea]**.

**Final dataset**

The final dataset consisted of 1,000 images divided in four categories, as follows: Drawings generated by Human Visual Artists: 283; Drawings generated by Human Non-Artists: 265; Drawings generated by HI-GenAI: 230; and Drawings generated by SG-GenAI: 222.

Images containing text, pseudo-text or colours were eliminated from the final dataset.

**Phase III: Creative Image Rating of Image Dataset by Human Raters**

To assess the creativity of the 1,000 generated images, an image rating task was designed and conducted through the online recruitment platform Prolific (See Figure 1). A total of 255 raters (Mean Age = 38.1 ±12.7 y.o.; 126 female, 124 male, 5 Other; Mean Years of Education = 15.9 ± 4 y.o.) took part and received an economical compensation for their participation.

Human raters were told that they would be shown a selection of 120 simple pencil drawings to be rated on five criteria. No mention was made of the different stimuli used to generate the drawings, nor of the origin of the drawings (Human and GenAI). Raters were then shown each drawing individually and asked to rate them using 7-point Likert scales from 1 (*Not at all*) to 7 (*Extremely*) across five dimensions: “Liking”, “Vividness”, “Originality”, “Aesthetics”, and “Curiosity & Interest”, presented in the same order for all raters (see Section S3).

To ensure an equal number of ratings for each of the 1,000 images, 250 unique subsets of 120 images each were generated. Each subset contained a balanced distribution of images from each of the four experimental categories, while also ensuring that each image would appear a minimum of 30 times across the 250 sets. During the rating task, the images were presented to participants in a randomized order.

The raters were first asked to complete the AReA questionnaire, a 14-item scale designed to assess an individual's responsiveness to aesthetic experiences of diverse nature (*40*). This allowed us to identify any potential impact of the individual raters’ aesthetic sensitivity on perceived creativity scores, and whether it led to different rating patterns emerging across each category of drawings.

**Phase IV: Image Dataset Rating by GenAI Rater**

A second rating of the image dataset was carried out using GPT-4o, in order to identify potential differences and similarities emerging between human and GenAI agents in their appraisal of visual creativity (See Figure 1). The GenAI rating of the database was carried out through the GPT-4o API, by simulating the conditions of the human image rating task through a script. The LLM was prompted to behave like a human rater taking part in an online image rating task, receiving identical scoring instructions to the human raters (see Section S3). Each image was run through the GPT-4o API 30 times, to obtain the same number of ratings as the human sample.

## Data analysis

The dataset comprised 60,600 ratings of the 1,000 drawings (30,600 by human participants and 30,000 by GPT-4o), containing, for each rating, the following information: the drawing's unique ID, the TCIA stimulus ID, the drawer type (Human or GenAI), the category (Visual artists, Non-artists, HI-GenAI and SG-GenAI), the rater's unique ID, the rater's type (Human or GPT-4o), and the scores on the five dimensions. For human raters, the dataset also included their age, gender, education level and AReA score. For human drawers, we also included the drawer's unique ID, age, gender, education, AReA score, and, for the artists, their field of practice and number of years of practice.

On this dataset, we applied various psychometric techniques (*66*) to the five dimensions and their scores. In particular, we employed *k*-means clustering of the scores, correlation matrices with hierarchical clustering of the dimensions, Mokken Scaling Analysis looking at the distribution of the Guttman errors and the homogeneity values *H* of the whole set and of each dimension, as well as the Automatic Item Selection Procedure [*aisp*], Principal Component Analysis (PCA) and exploratory and confirmatory Factor Analysis [EFA and CFA] using `R`'s (*67*) packages `factoextra`, `NbClust`, `mokken`, `psych` and `lavaan`. These seem to suggest that a one-factor model best fits the human raters, while a two-factor model might fit the GPT-4o data slightly better than the one-factor, but, combined with the fact that we only have five dimensions and one latent with only two problematic indicators, we decided to fit a one-factor model to these ratings as well. Nevertheless, fitting a one-factor model to both sets of ratings (using `lavaan`'s multiple groups) clearly shows that both the factor loadings and the latent means are significantly different, prompting us to estimate this latent Creativity score separately for the human and GPT-4o raters.

Third, we tested the effects of various predictors on this Creativity score using mixed effects Beta regressions (as these scores are, by definition, bounded) with the item and rater IDs as crossed random effects using `R`'s (*67*) `glmmTMB` package and their results were plotted using `sjPlot`.

## Authors Contribution

**SR**: Conceptualization; data curation; formal analysis; writing – original draft. **CAM**: Data curation; formal analysis; writing – original draft. **PA**: Conceptualization; writing – review and editing. **OP**: Formal analysis; writing – review and editing. **MPa**: Conceptualization; data curation; writing – review and editing. **MPe**: Conceptualization; writing – review and editing. **DD**: Formal analysis; writing – review and editing. **ARF**: Conceptualization; funding acquisition; project administration; resources; supervision; writing – review and editing. **XCC**: Conceptualization; data curation; funding acquisition; project administration; resources; supervision; writing – review and editing.

## Acknowledgements

We thank all participants for their valuable engagement in this study. We are also grateful to Arslan Javed for his insightful guidance on the use of the ChatGPT API, and to La Fura dels Baus – Fundació Èpica for providing an inspiring and stimulating environment.

## Funding Information

This project received funding from grant PID2023-151083NA-I00 funded by MICIU/AEI/10.13039/501100011033 and by ERDF/EU, grant PID2021-127130NB-100 funded by MICIU/AEI/10.13039/501100011033 and by ERDF/EU, the project Cátedra ENIA UAB-Cruïlla (TSI-100929-2023-2), and the European Researchers' Night and Researchers at Schools 2024-2025 (nº 101162558) by the European Performing Science Program (EPSP).

**SR** has been awarded with a Maria Capdevila Research Fellowship from the Computer Vision Center. **CAM** was supported by the Spanish Government with a predoctoral FPI fellowship (MCI22020), and **ARF** was supported by the FIAS fellowship Program, co-funded by the European Commission, Marie-Skłodowska-Curie Actions - COFUND Program, Grant n° 945408. **DD** acknowledges grant number PID2022-138501NB-I00 funded by MICIU/AEI/10.13039/501100011033 (Spain) and by ERDF/EU.

## Conflict of interest

The authors declare that they have no competing interests.

## Data availability

All data needed to evaluate the conclusions in the paper are present in the paper, the Supplementary Materials and/or the project's OSF.

The dataset can be provided by the corresponding author pending scientific review and a completed material transfer agreement.